\pdfoutput=1
\documentclass{INTERSPEECH2023}

\usepackage{siunitx}
\usepackage{tabularx}
\usepackage{multirow}
\usepackage{makecell}
\usepackage{hyperref}
\usepackage{amssymb}
\usepackage[table]{xcolor}

\expandafter\let\csname c@tblerows\endcsname\rownum

\newcolumntype{Z}{>{\centering\let\newline\\\arraybackslash\hspace{0pt}}X}
\definecolor{gray}{rgb}{0.9, 0.9, 0.9}

\makeatletter
\def\blfootnote{\xdef\@thefnmark{}\@footnotetext}
\makeatother
\interspeechcameraready

\title{Synthetic Cross-accent Data Augmentation for Automatic Speech Recognition}
\name{Philipp Klumpp$^{1,2}$, Pooja Chitkara$^1$, Leda Sar{\i}$^1$, Prashant Serai$^1$, Jilong Wu$^1$, Irina-Elena Veliche$^1$, Rongqing Huang$^1$, Qing He$^1$}
\address{
  $^1$Meta AI\\
  $^2$Pattern Recognition Lab, FAU Erlangen-N\"urnberg}
\email{philipp.klumpp@fau.de, ledasari@meta.com}

\begin{document}

\maketitle
 
\begin{abstract}
The awareness for biased ASR datasets or models has increased notably in recent years. Even for English, despite a vast amount of available training data, systems perform worse for non-native speakers.
\\In this work, we improve an accent-conversion model (ACM) which transforms native US-English speech into accented pronunciation. We include phonetic knowledge in the ACM training to provide accurate feedback about how well certain pronunciation patterns were recovered in the synthesized waveform. Furthermore, we investigate the feasibility of learned accent representations instead of static embeddings. Generated data was then used to train two state-of-the-art ASR systems.
\\We evaluated our approach on native and non-native English datasets and found that synthetically accented data helped the ASR to better understand speech from seen accents. This observation did not translate to unseen accents, and it was not observed for a model that had been pre-trained exclusively with native speech. 
\end{abstract}

\blfootnote{Work was done when Philipp was an intern at Meta AI\\Correspondence to Leda Sar{\i}: \href{mailto:ledasari@meta.com}{ledasari@meta.com}}

\section{Introduction}
Modern automatic speech recognition (ASR) models are trained with many thousands of hours of recorded speech samples. Whilst the ever increasing size of ASR datasets gives models a greater opportunity for exploration, it could easily introduce unrepresentative priors through the underlying data distribution. The open-sourced English subset of Common Voice~\cite{ardila2020common} for example features three times as many (self-reported) male speakers compared to females in its most recent version\footnote{\url{https://github.com/common-voice/cv-dataset}}.
\\Biased data itself may not necessarily result in a biased model. In a study investigating bias in ASR, a model trained with 1185 female and 1678 male Dutch speakers consistently produced lower test word error rates (WER) for the underrepresented group of female contributors~\cite{feng2021quantifying}. The same study also found differences in WER among different age groups which did not align with the age distribution of the training data. Whilst age and gender biases showed little manifestation in the final model performance, error rates increased significantly for non-native Dutch speakers. The same observation was made in a study about algorithmic bias in British English ASR~\cite{markl2022language}, highlighting that the patterns were not limited to second-language learner accents, but also observable among native speakers with different regional accents. Even the skin-tone, which might reflect a sociocultural background, could implicitly induce bias~\cite{liu2022towards}. The Artie Bias Corpus~\cite{meyer2020artie} provides a test dataset to evaluate to what degree biases, particularly those related to accented speech, would occur in an ASR system.
\\A poor representation of non-native compared to native speakers of a particular language hinders the speech recognition system to explore differences in pronunciation or language structure which originate from an underlying foreign native language.
\begin{figure}[t]
    \centering
    \includegraphics[width=1.0\linewidth]{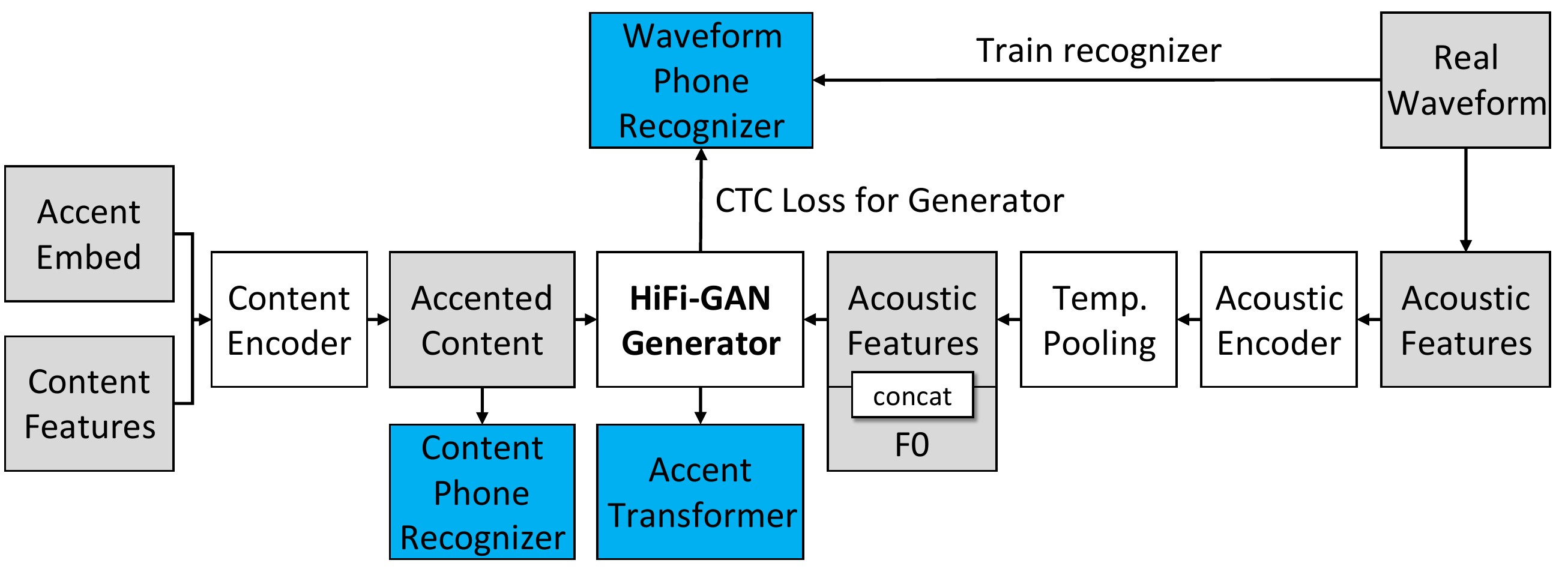}
    \caption{The accent conversion model uses a combination of content features resembling accent-specific pronunciation as well as a set of acoustic features to generate accent-converted speech. Blue blocks denote newly added modules.}
    \label{fig:acm_model}
\vspace{-6mm}
\end{figure}
For example, second-language learners, unlike infants, already acquired a phonetic inventory. The intersection of their native inventory and the inventory of the second language could introduce all kinds of pronunciation alternatives, which would result in an altered, but still intelligible utterance~\cite{flege1980phonetic}. The inability of ASR systems to maintain a high recognition quality for non-native speakers indicates that models which had been trained predominantly with samples from native speakers lack robustness against strongly underrepresented pronunciation patterns.
\\Providing an ASR with an auxiliary input describing the source accent helped improve the word error rate (WER) on accented speech~\cite{li2021accent}. The auxiliary input could be derived from learned embeddings, but also from an intermediate, unsupervised wav2vec representation~\cite{tjandra2022improved}. Besides accent embeddings, a multi-task learning scheme of speech recognition and accent classification has shown improvements over baseline systems~\cite{viglino2019end,jain2018improved}. By means of domain-adversarial training~\cite{sun2018domain}, AIPNet~\cite{chen2020aipnet} was able to learn accent-invariant acoustic feature vectors which could then be used by an ASR system to improve robustness on non-native speech samples.
\\Several approaches of accent conversion target the inverse problem of translating non-native accented speech to native accent, for example via linguistic and speaker representations in combination with a speech synthesis model conditioned on native speech~\cite{liu2020end}. In another approach, phonetic posteriorgrams of individual speakers were computed and then translated to a native accent domain to be used as input for a neural vocoder~\cite{zhao2018accent}. Conversion from native to foreign accent was achieved using a set of bottleneck features~\cite{zhang2022accentspeech} or accent-invariant representation of acoustic and content features~\cite{mumin2022acm}, where both works highlight the challenge of achieving good accent conversion while preserving the voice of the source speaker.
\\This work aims to convert speech samples from native American English to accented speech to synthetically generate foreign accent ASR training data. We trained two state-of-the-art speech recognizers with synthetic data and compared their performance to a baseline system that had not seen accented speech during optimization. Our main question was `Is it possible to improve the recognition of accented speech with synthetic samples of a particular accent?'. Furthermore, we evaluated the ability of generalization towards non-native pronunciation alternatives by training the ASR on a set of accents $A_{\text{train}}$ and test performance on another set $A_{\text{test}}$, such that $A_{\text{train}} \cap A_{\text{test}} \approx \varnothing$. Lastly, we evaluated the feasibility of learnable, non-static accent representations as they could help an ACM to derive pronunciation patterns for previously unseen accents.

\section{Methods}
We trained an ACM to convert native US English to speech of various foreign accents. The accent-converted speech data was then used to train more robust ASR models for non-native English. The ACM has been described in a previous work~\cite{mumin2022acm} and is outlined in Figure~\ref{fig:acm_model}. Its generative component is based on Hifi-GAN~\cite{kong2020hifi} and used a combination of acoustic features and content representations as input. The former should preserve the original voice, rhythm and intonation of the source speaker, whilst the latter combined a one-hot sequence of uttered characters predicted by a wav2vec~2.0 model~\cite{baevski2020wav2vec} with an accent-embedding to encode accent-specific pronunciation.
As in the original work~\cite{mumin2022acm}, the ACM was trained in an accent-symmetric fashion, meaning that the accent of the source (input) signal always matched the target accent of the generated waveform. This requirement was introduced by the reconstruction loss
\begin{equation}
    L_{mel} = ||M(\mathbf{x}) - M(\mathbf{\hat{x}})||_1
    \label{eq:l1_loss}
\end{equation}
where $M$ denotes the \SI{80}{}-dimensional mel spectrogram of the real ($\mathbf{x}$) and generated ($\mathbf{\hat{x}}$) waveforms. With matching source and target accents during training, accent-related information must be disentangled from the acoustic feature vector as this should originate exclusively from the content features.

\subsection{The Accent Transformer}
The accentedness of synthesized signals was promoted by introducing a sequence classification transformer module which predicted the accent of real and generated audio. It used \SI{13}{}~MFCC features and their first- and second-order derivatives as input, projected them to \SI{96}{}~channels per time step, which were then concatenated with a classification token. After a stack of \SI{10}{}~transformer encoder layers with feed-forward dimension of \SI{384}{} and a dropout of $p=0.1$, the output at the classification token was linearly projected to $N$ accents. This accent transformer (AT) was trained only on real audio samples. After warming up for \SI{20000}{}~steps, it was also shown generated waveforms, and the cross-entropy classification loss was propagated back into the generative model, providing feedback about how well a certain target accent was preserved in the generated speech signal. Because the AT was doing multi-accent rather than native versus non-native classification, we also adjusted the accent discriminator of the original ACM accordingly. Its purpose is to learn accent classification from acoustic features, and then applying an adversarial loss on said features to disentangle any accent information.

\subsection{Phonetic knowledge}
The proposed improvements of the ACM described in this work aim to strengthen the conversion of accent-specific pronunciations. The reconstruction error $L_{mel}$ in Equation~\ref{eq:l1_loss} had a strong contribution to the total generator loss. This promotes the synthesis quality of lower frequency bands, for which human beings have a higher perceptual resolution. Consequently, it will be less descriptive for many consonant phones, particularly unvoiced variants, as they manifest much more in higher frequencies. Another disadvantage of the reconstruction loss is the induced importance of phone duration. Longer phones, such as vowels, will inevitably experience a stronger optimization compared to that of short phones, like stops. To provide a more accurate feedback in terms of accented pronunciation quality, we included phonetic information in the ACM training. The ground truth information was derived from a machine-generated sequential annotation of training samples w.r.t. various phonetic features. They provided categorical description of the type (vowel, fricative, ...), position, first and second formant, duration, phonation, articulation, co-articulation and roundedness of every spoken phone. This allowed us to define a closed set of \SI{51}{}~phone classes, each resembling a unique combination of phonetic features. This phone sequence was used in two different ways. A pre-trained wav2vec~2.0 base model was fine-tuned to predict the phone class sequence of real audio samples using CTC loss~\cite{graves2012connectionist}. Just like the AT module, it was then used on the generated waveform and the gradients were back-propagated to the generator. The idea was to teach the phone recognizer what every phone should sound like from real waveforms, to then enforce equal pronunciation in the generated utterance.
\\Phonetic knowledge was also used to improve the quality of accented content features. Therefore, we added a second phone recognizer which would have to identify the same ground truth sequence as the wav2vec~2.0 module, but this time, the input was the content code. After applying a convolutional positional encoding, the result was fed to ten transformer layers with four attention heads and a feed-forward dimension of~\SI{128}{}. The output was once again projected to the \SI{51}{}~phone classes, plus a blank token, and optimized using  CTC loss.

\subsection{Learnable accent representations}
Using a static mapping, e.g. by means of a one-hot encoding, introduces certain limitations. The degree of accentedness in a speech signal is not controllable, hence it is not possible to take variable strengths of accent manifestation into account. Another downside is the supervised setup, limiting the ACM to the set of accents it had seen during training. Instead of a static embedding, it could also possible to learn an accent representation from the input signal. If it is successfully disentangled from all other information, such as speaker traits or acoustic environment, then this learned accent representation can be used to alter the accent of a secondary speech sample. Because the proposed ACM focused on the pronunciation characteristics of different accents, we used the average-pooled transformer output over time of the waveform-based phone recognizer, projected it to a lower-dimensional representation space, followed by another projection to $N$ accents. This representation block was trained for accent classification from the pooled transformer output, and the intermediate projection was utilized as the accent representation.

\subsection{Training setup}
The AT module was trained with the same hyper-parameters as all other discriminators of the original ACM~\cite{mumin2022acm}. The waveform- and content-based phone recognizers were trained using Adam optimizer ($\eta=3\cdot10^{-5}$) with exponential learning rate decay ($r=0.999995$) after each optimization step. The accent classification module for supervised representation learning was optimized with the same configuration as the AT. The entire ACM was trained with a batch-size of \SI{8}{}~samples of \SI{1}{s} length each.

\section{Experimental setup}
\subsection{ASR Models}
Two different model architectures were used to evaluate how synthetically accented speech could help improve ASR performance w.r.t. WER. An \SI{80}{}~bands mel spectrogram with \SI{10}{ms} frame-shift served as input. The spectrogram was augmented by masking out consecutive time steps and frequency bands as proposed in~\cite{park2019specaugment}. Both models employed an output vocabulary of \SI{5000}{}~sentence pieces~\cite{kudo2018sentencepiece} estimated over the \SI{960}{h} (clean-100 + clean-360 + other-500) training split of Librispeech~\cite{panayotov2015librispeech} (LS). The base model's (Base) encoder performed a linear projection of the \SI{80}{} mel bands to \SI{128}{} channels, followed by a time reduction step, merging every four consecutive time steps by concatenation along the feature dimension. Temporal context was obtained from a stack of \SI{20}{} efficient memory transformer (Emformer~\cite{shi2021emformer}) layers which employed \SI{8}{} attention heads, a feed-forward dimension of \SI{2048}{} and GELU activation~\cite{hendrycks2016gaussian}. The Emformer output was projected to \SI{1024}{} layer-normalized~\cite{ba2016layer} channels which then served as input for a recurrent neural transducer (RNNT). The RNNT predictor embedded all previously emitted symbols into a \SI{512}{}-dimensional space, using three LSTMs~\cite{hochreiter1997long}. The output of each recurrent layer was regularized by means of dropout~($p=0.3$) and lastly projected to \SI{1024}{} layer-normalized channels. The additive combination of encoder and predictor outputs was projected to a probability distribution over the output vocabulary plus a blank token.
\\The encoder structure of the HuBERT model was similar to that of the Base model, with slight modifications of channel configurations for certain layers. The dimensionality of the initial linear projection of mel-frequency bands was increased to \SI{192}{} channels. A subsequent residual 1D~convolution with \SI{31}{}~kernel elements was used as relative positional encoding. Due to the increased number of channels of the initial convolution, the time reduction layer's concatenation yielded a \SI{768}{}-dimensional input for the transformer blocks. No linear projection was applied to the last transformer layer's output (unlike for Base). The HuBERT model used the same transducer architecture as the Base model, with an increased dropout probability~($p=0.5$) after each recurrent layer and a higher dimensionality of the final projection (\SI{768}{}~channels) to match that of the predictor.
\\Both models were optimized by minimizing the transducer loss~\cite{graves2012sequence} using Adam optimizer~\cite{kingma2015adam} and weight decay, with the latter being significantly smaller for the pre-trained HuBERT model (Base:~$0.1$; HuBERT:~$10^{-6}$). We applied clipping of maximum gradient values to~$1.0$ and maximum gradient norm to~$10.0$.
\begin{table*}[t]
\small
    \centering
    \rowcolors{3}{}{gray}
    \begin{tabularx}{0.94\textwidth}{ll|cccc|ZZZ|Z}
       \multirow{2}{*}{\textbf{Experiment}} & \multirow{2}{*}{\textbf{Model}} & \multicolumn{2}{c}{\textbf{LS dev}} & \multicolumn{2}{c|}{\textbf{LS test}} & \multicolumn{3}{c|}{\textbf{L2-Arctic}} & \textbf{AVP} \\
      & & clean & other & clean & other & Arabic & Hindi & Mandarin & mean \\ \toprule
                & Base   & $3.8$    & $10.3$  & $4.1$   & $10.2$   & $24.7$  & $23.5$  & $33.9$  & $37.9$    \\ 
    \multirow{-2}{*}{\cellcolor[rgb]{1.,1.,1.}\makecell[l]{\textbf{LS only}}}
                & HuBERT & $9.5$    & $16.2$  & $9.8$   & $16.9$   & $26.9$  & $26.9$  & $34.2$  & $43.4$    \\ \midrule

                & Base   & $3.9$    & $10.4$  & $4.1$   & $10.4$   & $23.4$      & $23.3$    & $33.5$  & $38.4$   \\ 
    \multirow{-2}{*}{\cellcolor[rgb]{1.,1.,1.}\makecell[l]{\textbf{LS \& Arabic}}}
                & HuBERT & $9.3$    & $15.8$  & $9.6$   & $16.0$   & $31.2$      & $29.5$    & $35.4$  & $42.3$   \\ \midrule
                
                & Base   & $3.8$    & $10.3$  & $4.1$   & $10.3$   & $24.7$      & $22.0$    & $33.8$  & $38.3$   \\ 
    \multirow{-2}{*}{\cellcolor[rgb]{1.,1.,1.}\makecell[l]{\textbf{LS \& Hindi}}}
                & HuBERT & $9.3$    & $15.6$  & $9.7$   & $16.1$   & $26.2$      & $26.4$    & $33.4$  & $42.5$   \\ \midrule

                & Base   & $3.9$    & $10.4$  & $4.1$   & $10.5$   & $24.9$      & $23.5$    & $32.7$  & $38.9$   \\ 
    \multirow{-2}{*}{\cellcolor[rgb]{1.,1.,1.}\makecell[l]{\textbf{LS \& Mandarin}}}
                & HuBERT & $9.3$    & $15.7$  & $9.5$   & $15.9$   & $27.8$      & $30.0$    & $37.5$  & $42.7$   \\ \midrule
                
                & Base   & $3.6$    & $9.8$  & $3.8$   & $9.7$   & $23.4$  & $22.4$  & $32.2$  & $37.6$    \\ 
    \multirow{-2}{*}{\cellcolor[rgb]{1.,1.,1.}\makecell[l]{\textbf{LS \& 7 Accents}}}
                & HuBERT & $9.5$    & $15.8$  & $9.8$   & $16.1$   & $29.4$  & $28.0$  & $37.3$  & $43.1$    \\ \midrule
    \end{tabularx}
    \caption{WERs for US English (Librispeech), Arabic, Hindi \& Mandarin English was well as a collection of European accents (AVP). \textbf{LS only} is the baseline setup, where recognizers had only been trained with real US English data. \textbf{LS \& Accent} combines the original data with one synthetic accent. \textbf{LS \& 7 Accents} combines the original data with synthetic data of all seven non-US accents.}
    \label{tab:res_wer}
\vspace{-7mm}
\end{table*}
One batch of training samples would decode to a maximum of \SI{40000}{}~(Base) or \SI{5000}{}~(HuBERT) sentence piece tokens in the ground truth annotation. Because the Base model was trained from scratch with the entire LS-960h data, we increased the number of epochs to~\SI{180}{} compared to~\SI{120}{} for HuBERT fine-tuning. The learning rates of the Base ($10^{-3}$) and HuBERT ($10^{-4}$) model increased linearly during the first~\SI{10000}{} (HuBERT:~\SI{5000}{}) warm-up steps, then remained constant, and decayed exponentially by factor $0.96$ after \SI{60}{}~epochs.

\subsection{Datasets}
The data distribution and weighting of the original ACM was adopted: The L2-Arctic corpus~\cite{zhao2018l2} comprised English utterances from two female and two male speakers for Arabic, British, Hindi, Korean, Mandarin, Spanish and Vietnamese accent. The amount of Indian accent data was further increased by including Indic~TTS~\cite{indic_tts}, which contains two female and male speakers. For British English, we included the corresponding subset from the Speech Accent Archive~\cite{speech_accent_archive} in addition to the VCTK dataset~\cite{vc_for_accent_reduction}. For American English, the refined version of LS for speech synthesis applications was used (LibriTTS~\cite{zen2019libritts}), precisely the train-clean-100 and train-clean-360 splits.
\\The Base ASR model was optimized straight from initialization with LS~\SI{960}{h}. On the other hand, the HuBERT model had been pre-trained towards self-supervised HuBERT features~\cite{hsu2021hubert} using LS~\SI{960}{h}, and then fine-tuned only on the train-clean~\SI{100}{h} LS subset.
\\To enrich the described ASR training data with accented speech samples, we converted speech samples from LibriTTS train-clean-100 (US English) to the seven foreign accents. Notice that this procedure was never done during training of the ACM, in which the input accent always matched with the output accent. \SI{53}{h} of audio could be synthesized per accent. We chose to generate accented samples from said split because it is a subset of the original LS train-clean-100 split already used for training the ASR, hence it would not introduce new speakers or utterances.
\\The resulting ASR models were then tested on the clean and noisy development and test splits of LS. Furthermore, we evaluated ASR performance on three accents from the L2-Arctic corpus, as well as on the Accented Vox Populi (AVP) dataset~\cite{wang2021voxpopuli}. AVP is composed of speech samples recorded from events in the European Parliament, thus it includes many European accents.

\subsection{Experiments}
For the first experiment, we trained the baseline ASR systems without any accented speech data and evaluated their performance on the different splits from LS, the Arabic, Hindi and Mandarin subsets from L2-Arctic as well as on the different accents from AVP.
\begin{figure}[t]
    \centering
    \includegraphics[width=0.8\linewidth]{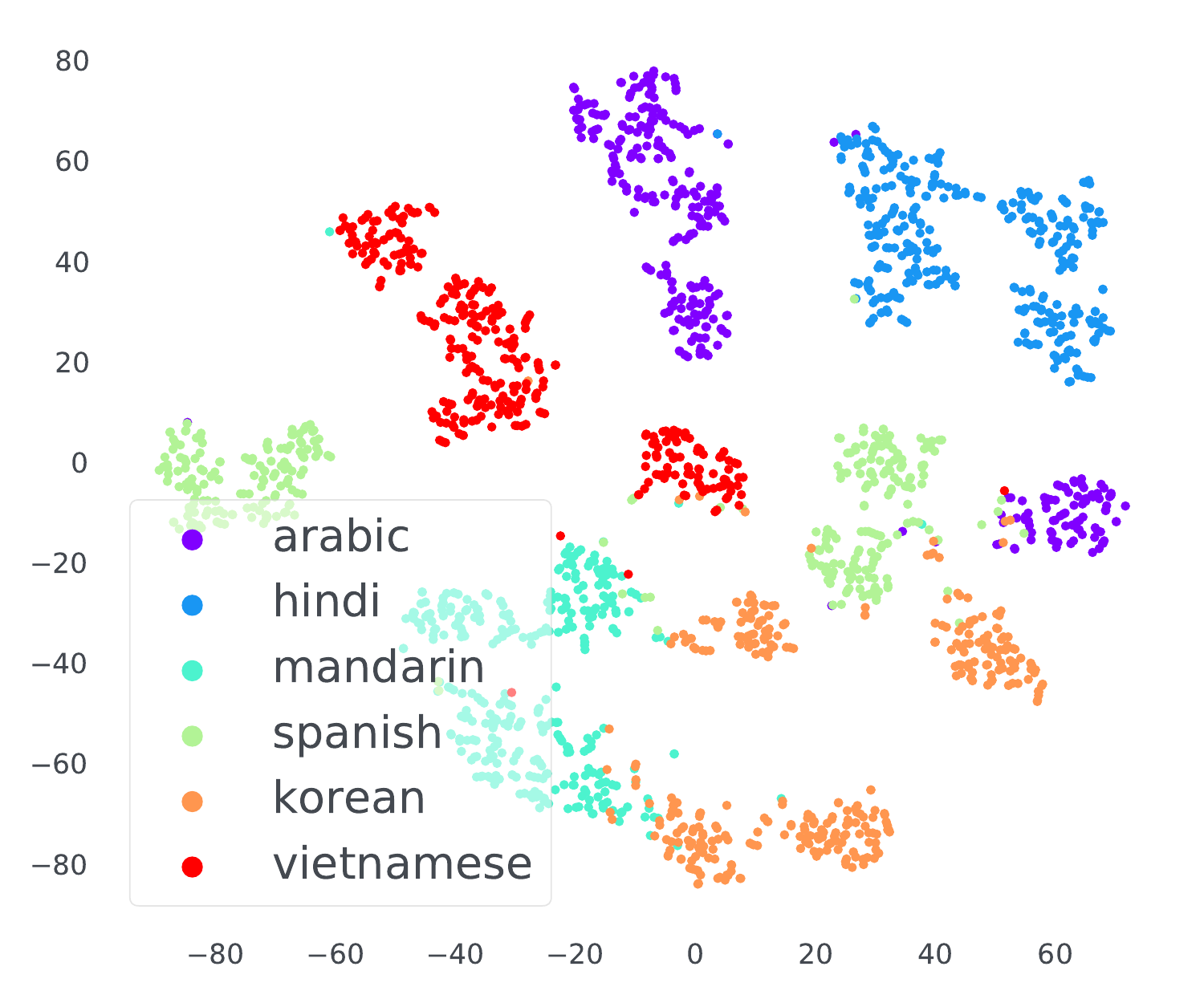}
    \caption{T-SNE embedding of learned accent representations from L2-Arctic samples. Some accent clusters are observable, but that was stongly superimposed by speaker information.}
    \label{fig:tsne_accents}
\vspace{-5mm}
\end{figure}
We then trained three ASR models with a combination of real training data plus \SI{53}{h} of synthetic accented data of the L2-Arctic accents used for ASR testing, applying a ratio of \SI{80}{\%} real and \SI{20}{\%} synthesized data.
\\Both ASR models were also optimized with the full set of all seven synthetic accents. The ratio of real and generated data remained the same as for the first experiment. 
\\In the last experiment, we visualized learned accent embeddings from L2-Arctic samples through a t-SNE plot to evaluate their suitability for encoding accent information in an ACM.

\section{Results}
Table~\ref{tab:res_wer} summarizes WER results for various experimental setups. WERs on LS were much lower compared to the other datasets, because the sentence piece lexicon had been estimated over this portion of data exclusively. Consequently, the ASR vocabulary would not be able to fully reconstruct samples from L2-Arctic or AVP.
\\As we included synthetically accented data for Arabic, Hindi or Mandarin in the ASR training, we found that WERs slightly improved for the particular accent on the Base model. For the HuBERT model however, we observed the opposite with a severe increase of WER for Arabic and Mandarin. Only Hindi improved by \SI{0.5}{\%}. At the same time, the additional accented data improved all WERs on the development and test portions from LS. On the european accents (AVP), the Base model always performed worse, HuBERT always performed better when using synthetic data.
\\After both ASR models had been trained with real data plus synthetic data from all 7~accents, we observed improving WERs for the noisy splits of LS. Only the Base model would also show minor improvements on the clean development and test splits. It also showed improvements on all accents of L2-Arctic, whilst the results remained almost the same for European accents (AVP). The HuBERT model continued to show worse results on the L2-Arctic data despite seeing accented speech during fine-tuning. Just like for the Base model, results remained almost unchanged on AVP.
\\Figure~\ref{fig:tsne_accents} visualizes t-SNE embedded accent representations learned from wav2vec 2.0 transformer outputs. Some accent clusters are clearly visible, but others appeared to show no coherence. Instead, they resembled each of the four speakers per accent in the L2-Arctic dataset.

\section{Discussion}
The inclusion of one synthetic accent during ASR training had a positive effect on recognition results for that particular accent, which was a clear indicator that the ACM was able to synthesize a sufficient degree of accentedness. The HuBERT model on the other hand suffered from synthetic data when testing on accented speech. A potential explanation is the lack of synthetic or accented data during pre-training. At the same time, fine-tuning HuBERT with synthetic data seemed to slightly improve robustness on noisy LS. The results on AVP showed little deviations, but due to the initially high WER values, there was apparently a large gap between the known LS vocabulary and that from AVP. None of the AVP results lead us to the conclusion that any model was able to learn general cross-accent robustness.
\\When all seven synthetic accents were included in training, the Base model would improve throughout all experiments. This could clearly be the result of an increased robustness to pronunciation alternatives. Again, this seemed not to be the case for unseen accents (AVP). The HuBERT variant, despite having a very similar architecture to the Base model, performed worse on accented speech, hence it could be helpful for such model to explore accented speech already during pre-training. As the Base model was optimized from scratch, it seemed to benefit much more from accented training samples.
\\We never utilized learned accent embeddings for sample generation throughout this study. Due to the very small number of unique speakers and the relatively large amount of audio for each contributor, the embeddings were not learning accent characteristic features in the first place, but rather classified the underlying speakers, hence they likely lack expressiveness for accent-specific pronunciation.

\section{Conclusion}
Synthetically accented speech samples can help improve the robustness of ASR systems against systematic pronunciation alternatives. This ability was not limited to a single accent, but was also observed when including multiple accents during training. However, these improvements did not manifest as a general accent robustness, as no noticeable progress could be reported for European accents. The learned accent representations seemed to encode accent, but were predominantly characterized by speaker clusters, indicating the need for a wider variety of speakers in the utilized accented speech data.

\bibliographystyle{IEEEtran}
\bibliography{template}

\end{document}